\documentclass[runningheads]{llncs}

\usepackage{eccv}

\usepackage{eccvabbrv}

\usepackage{graphicx}
\usepackage{booktabs}

\usepackage[accsupp]{axessibility}  

\usepackage[pagebackref,breaklinks,colorlinks,citecolor=eccvblue]{hyperref}

\usepackage{orcidlink}
\usepackage{multirow,booktabs}
\usepackage{amsmath}
\usepackage{amssymb}
\usepackage{color, colortbl}
\usepackage{times}
\usepackage{soul}
\usepackage{url}
\usepackage[vlined, ruled,boxed,linesnumbered]{algorithm2e}

\definecolor{Gray}{gray}{0.9}
\definecolor{dark_green}{HTML}{82B366}
\definecolor{dark_red}{HTML}{B85450}

\begin{document}

\title{Contrastive Adversarial Training for Unsupervised Domain Adaptation} 

\titlerunning{CAT}

\author{Jiahong Chen\inst{1} \and
Zhilin Zhang\inst{1} \and
Xin Li\inst{1}\and
Behzad Shahrasbi\inst{1}\and
Arjun Mishra\inst{1}
} 

\authorrunning{J.~Chen et al.}

\institute{Amazon\\
\email{\{jiahoc, zzhilin, lilc, behzadsb, misarjun\}@amazon.com}}

\maketitle

\begin{abstract}
    Domain adversarial training has shown its effective capability for finding domain invariant feature representations and been successfully adopted for various domain adaptation tasks. However, recent advances of large models (e.g., vision transformers) and emerging of complex adaptation scenarios (e.g., DomainNet) make adversarial training being easily biased towards source domain and hardly adapted to target domain. The reason is twofold: relying on large amount of labelled data from source domain for large model training and lacking of labelled data from target domain for fine-tuning. Existing approaches widely focused on either enhancing discriminator or improving the training stability for the backbone networks. Due to unbalanced competition between the feature extractor and the discriminator during the adversarial training, existing solutions fail to function well on complex datasets. To address this issue, we proposed a novel contrastive adversarial training (CAT) approach that leverages the labeled source domain samples to reinforce and regulate the feature generation for target domain. Typically, the regulation forces the target feature distribution being similar to the source feature distribution. CAT addressed three major challenges in adversarial learning: 1) ensure the feature distributions from two domains as indistinguishable as possible for the discriminator, resulting in a more robust domain-invariant feature generation; 2) encourage target samples moving closer to the source in the feature space, reducing the requirement for generalizing classifier trained on the labeled source domain to unlabeled target domain; 3) avoid directly aligning unpaired source and target samples within mini-batch. CAT can be easily plugged into existing models and exhibits significant performance improvements.
    We conduct extensive experiments on large domain adaptation benchmark datasets, and the results show that CAT significantly improves the baseline adversarial training methods from +0.5\% to +4.4\% on Visda-2017, from +1.0\% to +2.7\% on DomainNet, and from +0.3\% to +1.8\% on Office-Home, respectively. 
    \keywords{Domain Adaptation \and Adversarial Learning}
\end{abstract}

\section{Introduction}
Deep neural networks have achieved impressive performance in various computer vision tasks. However, such success often relies on a large amount of labeled training data, if we directly adopt large pretrained model for individual tasks, the performance will be degraded due to mis-alignment between source and target domain. On the other hand, it is infeasible and costly to obtain labelled data for training dedicated large models in most real-world scenarios. Unsupervised Domain Adaptation (UDA) addressed the label missing challenges by transferring knowledge from a label-rich source domain to a separate unlabeled target domain \cite{ben2006analysis}. Over the past years, many adversarial UDA adaptation methods have been proposed, which commonly leverages the idea of adversarial training to learn the domain-invariant feature representation by using a domain discriminator to compete with feature generator \cite{ganin2016domain,long2018conditional,rangwani2022closer}. These UDA methods are usually applied in conjunction with a pretrained Convolutional Neural Network (CNN) backbone (e.g., ResNet \cite{he2016deep}) and work well for small to medium sized images classification tasks, such as Office-Home \cite{venkateswara2017deep} and VisDA \cite{peng2017visda}. However, the performance of CNN-based adversarial approach on large-scale datasets like DomainNet \cite{peng2019moment} is unfavorable. To address such issue, recent researches have widely leverage the vision transformers (ViT) to resolve the challenging cross-domain adaption problems, i.e., unsupervised domain adaptation (UDA) task. 
Due to the success of transformer architecture, ViT-based approaches shown substantial improvements over the CNN-based frameworks \cite{xu2021cdtrans,yang2023tvt,zhu2023patch}. 



The above existing adversarial training methods widely consist of two parts: 1) a standard feed-forward neural network (usually including a feature generator and a label predictor, i.e., classifier); and 2) a discriminator. The discriminator connects to the feature extractor via a gradient reversal layer (GRL). During the back-propagation, GRL forces the feature distributions over the two domains to be similar through multiplying the gradients by a certain negative constant. In this way, source and target domains become indistinguishable for the discriminator. On the other hand, the classifier forces the feature generator to extract robust discriminative features for the labeled source domain. Through such adversarial learning, the discriminator competes with the feature generator for extracting the domain-invariant features across the two domains. It, therefore, enables a smooth knowledge adaptation from source domain to target domain.

Considering both backbone networks and adaptation scenarios become more complex over time, traditional adversarial training methods becomes less effective. Researchers have invested a lot of efforts in improving the domain adversarial training by improving discriminator \cite{long2018conditional}, or improving the training stability for the backbone network \cite{sun2022safe,rangwani2022closer}. However, existing adversarial training approaches purely relies on GRLs to force source and target domains appearing similar in order to better compete with discriminator. It limits the capability of adversarial training between the feature extractor and the discriminator. Typically, the improvement of the feature extractor will lead to further biased feature generation toward labeled source data, which in the meantime, makes the features from two data domains dis-similar and makes the domain discrimination takes become easier. As a result, it degrades the performance for target domain classification. Additionally, due to random sampling, the source and target samples are unpaired within each mini-batch. Without the knowledge of target labels, it is infeasible to directly align source and target features to improve the difficulty for discriminator \cite{wang2021discriminative}. Therefore, as more powerful backbone networks and discriminators are developed, the balanced competition between the feature extractor and the discriminator during the adversarial training is easily destroyed.



\begin{figure*}
\centering
\begin{subfigure}{.5\textwidth}
  \centering
  \includegraphics[width=.9\linewidth]{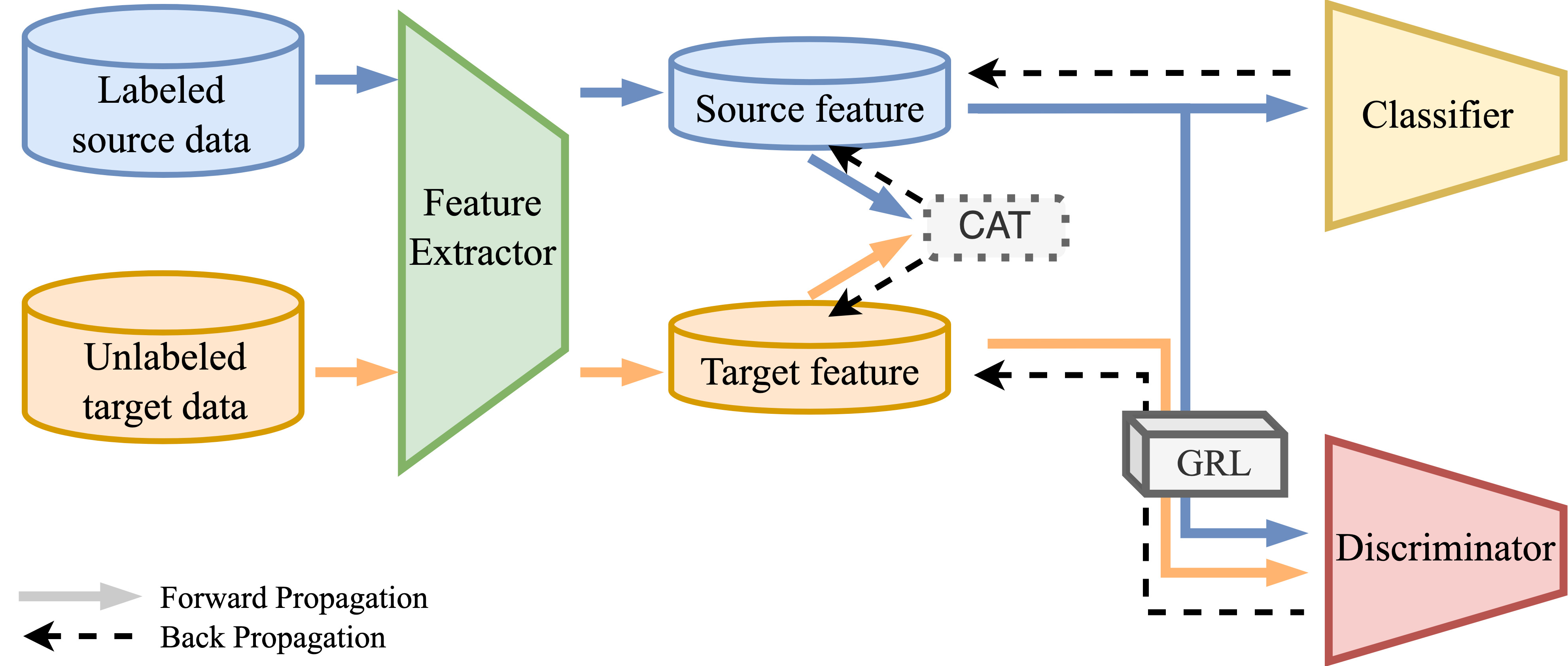}
  \caption{}
\end{subfigure}%
\begin{subfigure}{.5\textwidth}
  \centering
  \includegraphics[width=.8\linewidth]{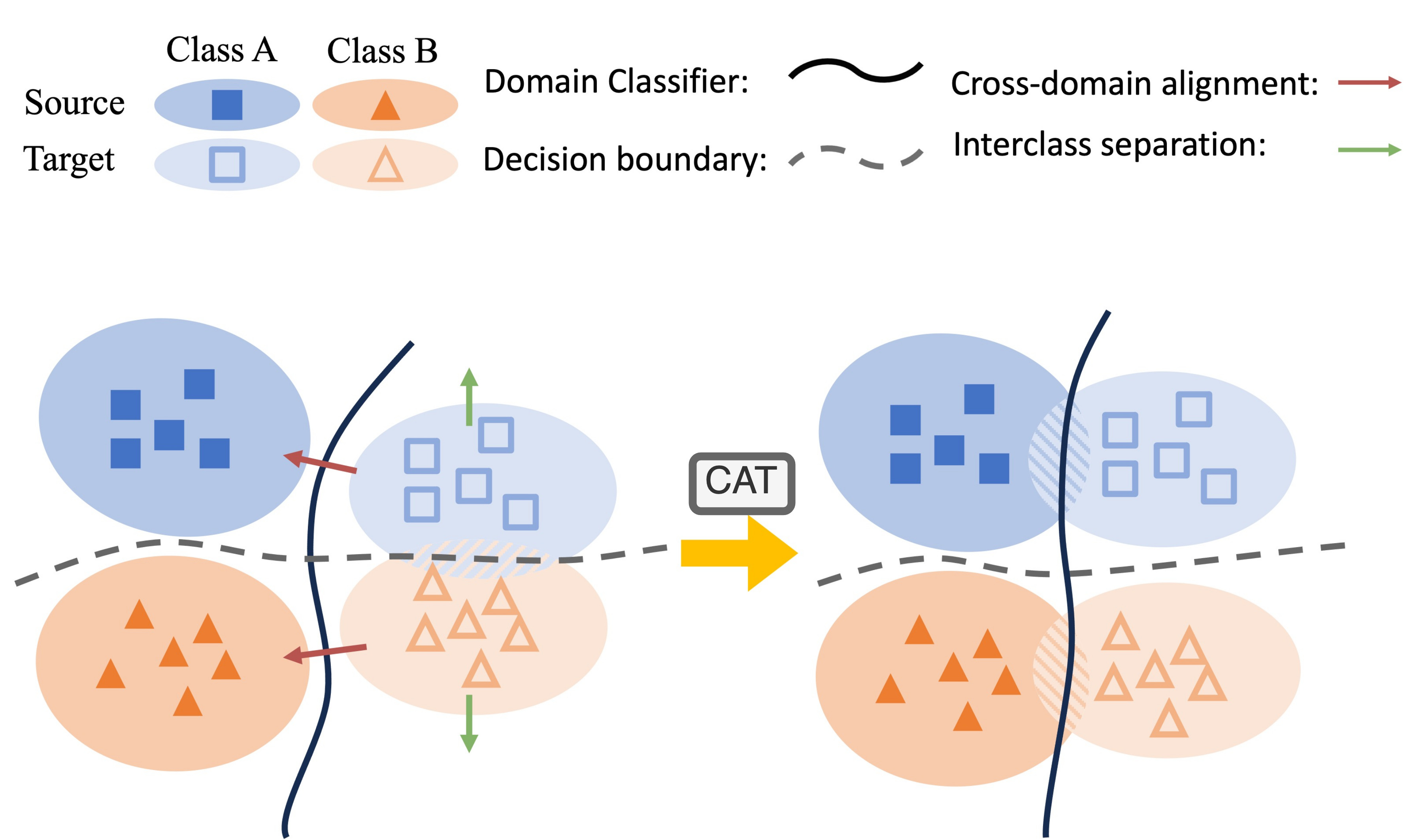}
  \caption{}
\end{subfigure}
\caption{Overview of Contrastive Adversarial Training (CAT). (a) highlights the location of our CAT component in the model, which serves as a reinforcement for the GRL process to make the domain classification more difficult for the discriminator; (b) illustrates that contrastively aligning target samples to it's source anchor can simultaneously increase the competitiveness of the adversarial training  for a more robust domain invariant feature extraction and reduce the domain discrepancy to more easily draw decision boundaries for unlabeled target domain.
}
\label{fig:overview}
\end{figure*}

To tackle such imbalanced competition between the feature extractor and the discriminator, we propose a novel Contrastive Adversarial Training (CAT) plug-in for adversarial UDAs (see Fig. \ref{fig:overview} (a)). CAT explicitly forces the features from two data domain become similar to increase the difficulty of the discriminator for distinguishing the samples from two different domains, while perverse the performance of the source classifier. Specifically, CAT mitigates the divergence between source and target in a contrastive way: source domain samples serves as the anchors, while similar target domain samples are pulled closer to the anchors and dissimilar target samples are pushed away. Considering that there are no ground truth labels for the target domain samples to align with anchors, we leverage a feature bank to find the similar and dissimilar target samples for the source domain samples at the global scale by gradually updating the feature bank with newly seen samples in each mini-batch. To this end, the target samples and the source samples from the same class are are clustered together, which leads to two benefits as shown in Fig. \ref{fig:overview} (b): 1) The decision boundaries learnt from the source domain can also be easily applied to the target due to the target-to-source alignment (intraclass alignment); 2) It becomes more difficult for the discriminator to tell the original domain of the samples due to the cross-domain interclass alignment, improving the robustness for both feature extractor and discriminator during the adversarial training. 


To the best of our knowledge, CAT is the first adversrial UDA approach to leverage the Contrastive Learning process between source and target domain for facilitating a more effective adversarial learning. Due to its universality and simplicity, CAT can be straightforwardly integrated into adversarial UDA methods across different datasets, making it highly accessible in practice. Additionally, we extensively verify the empirical efficacy of CAT across large UDA datasets for image classification with Vision Transformer (ViT) backbones \cite{dosovitskiy2020vit}. In summary, we make the following contributions: 

\begin{itemize}
    \item We uncover that the adversarial training for UDA part needs to be balanced by reinforcing GRL to compete with discriminator, leading to a more competitive process for adversrial training.
    \item We propose a novel contrastive loss that avoids the unpair problem in domain alignment. It uses the labeled source domain samples as anchor that pulls similar samples in target domain while pushing dis-similar target samples away, ensuring a easy decision boundary and a hard domain classification.
    \item The proposed model, CAT, can serve as a plug-in for adversarial UDAs and be integrated into existing adversarial training models easily.
    \item Extensive experiments are conducted on widely tested benchmarks. Our proposed CAT achieves the best performances, including 93.3\% on VisDA-2017, 85.1\% on Office-Home, and 58.0\% on DomainNet. It improves performance of the adversarial UDA models from +0.5\% to +4.4\% on Visda-2017, from +1.0\% to +2.7\% on DomainNet, and from +0.3\% to +1.8\% on Office-Home, respectively. 
\end{itemize}


\section{Related Work}
\textbf{Unsupervised Domain Adaptation:}
UDA has been proposed to address the challenges in learning the feature representations from the source domain’s labeled data and generalizes well on the unseen data from the target domain \cite{jin2020minimum,hoyer2023mic}. Ganin et al. first proposed Domain-Adversarial Training of Neural Networks (DANN), which leverage the power of adversarial training to address the cross-domain discrepancy \cite{ganin2016domain}. Then, Conditional Adversarial Domain Adaptation (CDAN) is proposed to improve the discriminator with conditional GAN \cite{long2018conditional}. Adversarial Discriminative Domain Adaptation (ADDA) is proposed to learn the mapping between source and target mappings to minimize the domain divergence, making the source classifier can be directly applied to the target feature \cite{tzeng2017adversarial}. 
Additionally, multiple methods focusing on improving the training stability for the backbone network are developed. Smooth Domain Adversarial Training (SDAT) addresses the problem of sub-optimal performance by leveraging the smoothness enhancing formulation for adversarial loss \cite{rangwani2022closer}. SSRT \cite{sun2022safe} adopts a \textit{safe training} \cite{li2014towards} strategy to reduce the risk of model collapse and improve the effectiveness of knowledge transfer between domains with large gaps.

\textbf{Contrastive Learning:} Contrastive Learning (CL) is a popular technique for self-supervised learning of visual representations \cite{ho2020contrastive,chen2020simple}. It uses pairs of examples to extract meaningful representations by contrasting positive and negative pairs of instances by leveraging the assumption that similar samples should be in close proximity within the learned feature space, while dissimilar instances should be positioned farther apart \cite{yang2022attracting}. CL has also been used in non-adversarial UDAs. Contrastive Adaptation Network (CAN) is a non-adversarial UDA approach that explicitly models the intraclass and interclass domain discrepancy \cite{kang2019contrastive}. Similarly, cross domain contrastive learning (CDCL) \cite{wang2022cross} further improves the contrastive learning for UDA with InfoNCE \cite{oord2018representation,he2020momentum}, which minimizes the distance of a positive pair relative to all other pairs \cite{sohn2016improved}. However, both CAN and CDCL focus on non-adversarial setting, which helps the classifier to accurately draw decision boundaries for unlabeled target domain, but ignored the effectiveness of incorporating adversarial training to further extract domain invariant loss.

\section{Methods}
\subsection{Problem Formulation for UDA}
The general setting of UDA consists of a labeled source domain $D_s = \{(x_s^{(i)} , y_s^{(i)})\}^N_{i=1}$ and a unlabeled target domain with $D_t = \{x_t^{(j)}\}^M_{j=1}$, where $x_{\cdot}$, $y_{\cdot}$ denotes data samples and labels respectively; $M$, $N$ represents the number of samples in each domain.
UDA aims to learn a model $h = g\circ f$, where $g (\cdot; \theta_g ): X \rightarrow Z$ denotes the feature extractor that projects the input samples into its feature space, $f(\cdot; \theta_f ) : Z \rightarrow Y$ denotes the classifier that projects the features $Z$ into logits. In the mean time, an adversarial adaptation learns domain-invariant feature via a binary domain discrimination $d(·; \theta_d) : Z \rightarrow \{0, 1\}$ that maps features to domain labels. 

The main learning objective for adversarial UDA consists of standard source classification loss $\mathcal{L}_{cls}$ and and domain adversarial loss $\mathcal{L}_d$:
\begin{equation}
\mathcal{L}_{cls} = -\mathbb{E}_{(\mathbf{x}_s , \mathbf{y}_s)\in D_s}L(f(g(\mathbf{x}_s)), \mathbf{y}_s),
\end{equation}
\begin{equation}
    \mathcal{L}_d = -\mathbb{E}_{\mathbf{x}_s\sim D_s} [\log d(f (\mathbf{x}_s))]-\mathbb{E}_{\mathbf{x}_t\sim D_t} [\log(1 - d(f(\mathbf{x}_t)))].
\end{equation}
where $L$ is the standard cross-entropy loss for classification that guarantee lower source risk. The overall objective function serving as the backbone for adversarial training is as follows:
\begin{equation}
    \min \mathcal{L} = \min \mathcal{L}_{cls} + \mathcal{L}_d.
\end{equation}


\subsection{Contrastive Source-Target Alignment}\label{sec:neighbor-finding}
To explicitly encourage closer distance for two data domains in feature spaces, we propose to mitigate the divergence between source and target in a contrastive way. Typically, source domain samples act as the anchors, followed by pulling similar target domain samples towards them and pushing dissimilar ones away. To facilitate this process, we introduce feature clustering, aiming to group highly related target features towards the source samples while concurrently creating separation among dissimilar ones. 

In order to calculate the sample-wise similarity, our approach draws inspiration from the principles of unsupervised representation learning's neighborhood discovery, as exemplified in prior works such as \cite{saito2020universal}, \cite{van2020scan}, and \cite{yang2022attracting}. To keep track of unlabeled target domain features, we build a feature bank ($\mathcal{B}$) to help the contrastive learning process. Typically, we collect the features of the target domain samples, $\mathcal{B} = \{ \mathbf{z}_t = g(\mathbf{x}_t) | \mathbf{x}_t \in D_t \}$, and continuously update  $\mathcal{B}$ during training. This iterative update mechanism allows all target domain samples' features to be continuously improved throughout the training process. 

For each source sample $\mathbf{x}^{(i)}_s$ in the mini-batch, we calculate the similarity, $p_{ij}$, between its source feature $\mathbf{z}^{(i)}_s$ and the target feature $\mathbf{z}^{(j)}_t$ of a target sample $\mathbf{x}^{(j)}_t$ in terms of softmax function to captures how much a given anchor source feature $\mathbf{z}^{(i)}_s$ aligns with the target feature $\mathbf{z}^{(j)}_t$
\begin{equation}\label{eqn:prob_func}
p_{ij} = \frac{e^{\mathbf{z}^{(i)}_s\cdot \mathbf{z}^{(j)}_t}}{\sum_{k=1}^{M} e^{\mathbf{z}^{(i)}_s \cdot  \mathbf{z}^{(k)}_t}}.
\end{equation}

Equation \ref{eqn:prob_func} can be interpreted as the probability indicating the similarity between a target feature $\mathbf{z}^{(j)}_t$ is compared to a source feature $\mathbf{z}^{(i)}_s$. Then, the likelihood of source-target similarity can be denoted as the product of probabilities, following the work of Yang et al. \cite{yang2022attracting}:
\begin{equation}
\begin{aligned}
P(\mathbf{z_s}^{(i)}|\mathcal{C}_{i}, \theta_g) =\prod_{j \in \mathcal{C}_i} p_{ij},~P(\mathbf{z_s}^{(i)}|\mathcal{D}_{i}, \theta_g) =\prod_{j \in \mathcal{D}_i} p_{ij} 
\end{aligned}
\end{equation}
where $\mathcal{C}_i$ is the close set for source anchor $\mathbf{z}^{(i)}_s$ and $\mathcal{D}_i$ denotes the distant set. The close set $\mathcal{C}_i$ consists of features of the target domain's samples that are considered close to source anchor $\mathbf{z}^{(i)}_s$, while the distant set $\mathcal{D}_i$ includes features of target domain samples that are considered distant from the anchor. To define the neighbor set $\mathcal{C}_i(\mathbf{z}^{(i)}_s)$ for a query sample $\mathbf{x}^{(i)}_s$, we leverage the concept of $K$-Nearest Neighbors ($K$NN) \cite{cover1967nearest} from the target domain based on the cosine similarity in the feature space:
\begin{equation}
\mathcal{C}_{i} = \{\text{argmin}_{n=1, g(\mathbf{x}^{(n)}_t)\in \mathcal{B}}^{K} [g(\mathbf{x}^{(i)}_s) \cdot g(\mathbf{x}^{(n)}_t])\}.
\end{equation}

Similarly, the distant set $\mathcal{D}_i(\mathbf{z}^{(i)}_s)$ for a query sample $\mathbf{x}^{(i)}_s$ is defined by its $K$-Farthest Neighbors from the target domain based on the cosine similarity, as shown below:

\begin{equation}
\mathcal{D}_{i}  = \{\text{argmax}_{m=1, g(\mathbf{x}^{(m)}_t)\in \mathcal{B}}^{K} [g(\mathbf{x}^{(i)}_s) \cdot g(\mathbf{x}^{(m)}_t)]\}.
\end{equation}

%




Intuitively, for each source sample $\mathbf{x_s}^{(i)}$, the corresponding target features in $\mathcal{D}_i $ should be less similar to its feature $\mathbf{z_s}^{(i)}$ than those in $\mathcal{C}_i $. This process involves moving target features in $\mathcal{C}_i$ towards $\mathbf{z}^{(i)}_s$ while pushing away target features in $\mathcal{D}_i$, enabling contrastive learning to identify and cluster discriminative features from unlabeled data within the target domain. Our goal is to maximize the likelihood of target feature clustering in a contrastive manner by simultaneously maximizing the likelihood of the neighbor set and minimizing the likelihood of the distant set. The optimization objective can be formulated as follows:


\begin{equation}\label{eqn:max-likelihood}
    \max \log\left[\frac{P(\mathbf{z_s}^{(i)}|\mathcal{C}_i, \theta_g)}{P(\mathbf{z_s}^{(i)}|\mathcal{D}_i, \theta_g)}\right],
\end{equation}
which can be equivalently expressed as the minimization of the product of probabilities in the away set $\mathcal{D}_i$ divided by the product of probabilities in the neighbor set:
\begin{equation}\label{eqn:log-expansion}
\begin{aligned}
    &\max \log\left[\frac{\prod_{n \in \mathcal{C}_i} p_{i n}}{\prod_{m \in \mathcal{D}_i} p_{i m}}\right].\\
    =& \max\left[\log\prod_{n \in \mathcal{C}_i} p_{i n}-\log\prod_{m \in \mathcal{D}_i} p_{i m}\right]\\
    =&\max \left[\sum_{n \in \mathcal{C}_i} \log (p_{i n})-\sum_{m \in \mathcal{D}_i} \log (p_{i m})\right]\\
\end{aligned}
\end{equation}

We can then rewrite Eqn. \ref{eqn:log-expansion} using Eqn. \ref{eqn:prob_func}:
\begin{equation}\label{eqn:expand-prob}
\begin{aligned}
    &\max \left[\sum_{n \in \mathcal{C}_i} \log (\frac{e^{\mathbf{z}^{(i)}_s\cdot \mathbf{z}^{(n)}_t}}{\sum_{k=1}^M e^{\mathbf{z}^{(i)}_s \cdot  \mathbf{z}^{(k)}_t}})-\sum_{m \in \mathcal{D}_i} \log (\frac{e^{\mathbf{z}^{(i)}_s\cdot \mathbf{z}^{(m)}_t}}{\sum_{k=1}^M e^{\mathbf{z}^{(i)}_s \cdot  \mathbf{z}^{(k)}_t}})\right]\\
    =&\max \left[\sum_{n \in \mathcal{C}_i} \mathbf{z}^{(i)}_s\cdot \mathbf{z}^{(n)}_t-\sum_{m \in \mathcal{D}_i} \mathbf{z}^{(i)}_s\cdot \mathbf{z}^{(m)}_t\right].
\end{aligned}
\end{equation}

Maximizing Eqn. \ref{eqn:expand-prob} can be achieved by the optimization in a contrastive way, i.e., maximizing dot production between source anchor and target features in the close set, while minimizing the ones for the distant set. Further expanding Eqn. \ref{eqn:expand-prob} can derive it in the form of triplet contrastive loss \cite{balntas2016learning}:
\begin{equation}\label{eqn:triplet}
\begin{aligned}
    &\max \left[\sum_{n \in \mathcal{C}_i} \mathbf{z}^{(i)}_s\cdot \mathbf{z}^{(n)}_t-\sum_{m \in \mathcal{D}_i} \mathbf{z}^{(i)}_s\cdot \mathbf{z}^{(m)}_t\right]\\
    \propto & \min \left[2\cdot \sum_{m \in \mathcal{D}_i} \mathbf{z}^{(i)}_s\cdot \mathbf{z}^{(m)}_t - 2\cdot \sum_{n \in \mathcal{C}_i} \mathbf{z}^{(i)}_s\cdot \mathbf{z}^{(n)}_t \right]\\
    \leq &\min \left[ \sum_{n \in \mathcal{C}_i} ||\mathbf{z}^{(i)}_s-\mathbf{z}^{(n)}_t||_2 - \sum_{m \in \mathcal{D}_i} ||\mathbf{z}^{(i)}_s-\mathbf{z}^{(m)}_t||_2\right],
\end{aligned}
\end{equation}
where $||\mathbf{z}^{(i)}_s-\mathbf{z}^{(n)}_t||_2$ can be regarded as the $l_2$-norm between the source anchor and target features in the close set $\mathcal{C}_i$, $||\mathbf{z}^{(i)}_s-\mathbf{z}^{(m)}_t||_2$ is the $l_2$-norm between the source anchor and target features in the distant set $\mathcal{D}_i$. Therefore, minimizing the loss function in Eqn. \ref{eqn:triplet} will pull the similar target features in the close set towards the source anchor, while pushing dis-similar target features in the distant set away from the anchor.

Additionally, optimizing Eqn. \ref{eqn:triplet} will result in a lower bound of Eqn. \ref{eqn:max-likelihood}, revealing that the maximum likelihood can be achieved by simultaneously minimizing the loss among source anchor, close target features, and distant target features.

This optimization objective effectively encourages the contrastive model's feature representation, learned through the feature generator $g$, to bring target samples in the same class as the source anchor closer to each other in the feature space, while simultaneously distancing dis-similar target samples of different classes. To this end, the source and target features become more distinguishable, making it 1) harder for the discriminator to identify the domain of input samples; 2) easier for the classifier to be applied in the unlabeled target features.


The contrastive loss from Equation \ref{eqn:triplet} can be regarded as the backbone for our new contrastive adversarial training. It can be directly integrated into other adversarial training methods such as CDAN, SDAT, SSRT. The detailed implementation for the proposed adversarial training plug-in, CAT, is summarized in Algorithm \ref{alg:CAT}.
\begin{algorithm}[h]
   \caption{The implementation of contrastive adversarial training for CAT}
   \label{alg:CAT}
  \SetAlgoLined
\KwIn{data samples/labels $D_s, D_t$, neighborhood size $K$, batch size $N'$, hyper-parameter $\gamma$, network structures $g$, $f$, coefficient $\lambda$.}
\KwOut{network parameters $\theta_g, \theta_f$.}
Initialize network parameters $\theta_g, \theta_f$\;
\For{\textup{sampled minibatch } $\mathbf{X}_s, \mathbf{Y}_s, \mathbf{X}_t$}{
\textcolor{gray}{\# get model outputs}\\
$\mathbf{Z}_s \leftarrow g(\mathbf{X}_s;\theta_g), \mathbf{Z}_t \leftarrow g(\mathbf{X}_t;\theta_g)$\;
$\hat{\mathbf{Y}}_s \leftarrow f(\mathbf{Z}_s; \theta_f)$\;
\textcolor{gray}{\# pairwise similarity}\\
\For{$i \in \{1, \cdots, N\}, j\in \{1, \dots, M\}$}{
    $D_{cos}(\mathbf{z}^{(i)}_s, \mathbf{z}^{(j)}_t) = \mathbf{z}^{(i)}_s \cdot \mathbf{z}^{(j)}_t/||\mathbf{z}^{(i)}_s||\cdot||\mathbf{z}^{(j)}_t||$
}
\textcolor{gray}{\# generate neighbor set and away set}\\
$\mathcal{C}_i = \{\text{argmin}_{j=1}^{K} [D_{cos}(\mathbf{z}^{(i)}_s, \mathbf{z}^{(j)}_t)]\}$\;
$\mathcal{D}_i = \{\text{argmax}_{j=1}^{K} [D_{cos}(\mathbf{z}^{(i)}_s, \mathbf{z}^{(j)}_t)]\}$\;
\textcolor{gray}{\# calculate contrastive loss}\\
$\mathcal{L}_{con} = \frac{1}{N'} \sum\limits_{i=1}^{N'} \left[\sum\limits_{n \in \mathcal{C}_i} ||\mathbf{z}^{(i)}_s-\mathbf{z}^{(n)}_t||_2-\sum\limits_{m \in \mathcal{D}_i} ||\mathbf{z}^{(i)}_s-\mathbf{z}^{(m)}_t||_2\right]$\; 
\textcolor{gray}{\# calculate overall loss}\\
$\mathcal{L} = \mathcal{L}_{cls}+\mathcal{L}_{d}+\lambda\mathcal{L}_{con}$\;
Update model parameters with GRL $\rightarrow \theta_g, \theta_f$.\\
}
\end{algorithm}

\section{Experiments}
We evaluate our proposed method, CAT, on three datasets: DomainNet, VisDA-2017, and Office-Home, by integrating it to several state-of-the art algorithms with ViT backbone. The experiments were ran on AWS SageMaker P3 instances with NVIDIA Tesla V100 GPUs. All experiments were conducted 5 times and the average accuracy is reported. 
\subsection{Datasets}
~~~~\textbf{DomainNet} \cite{peng2019moment}: DomainNet is the largest dataset to date for domain adaptation, which consists of ~600K images for 345 different classes. In the experiment, we follow the set up in SDAT to test 20 different adaptation scenarios across 5 different domains: clipart (clp), infograph (inf), painting (pnt), real (rel), and sketch (skt).

\textbf{VisDA-2017} \cite{visda2017}: VisDA is a large domain adaptation dataset for the transition from simulation to real world, which contains 280K images for 12 different classes.

\textbf{Office-Home} \cite{venkateswara2017deep}: Office-Home is a common benchmark dataset for domain adaptation which contains 15,500 images from 65 classes and 4 domains: Art (A), Clipart (C), Product (Pr) and Real World (R).

\subsection{Implementation Details}
For UDA training, we follow SDAT and MIC \cite{rangwani2022closer,hoyer2023mic} by adopting the Transfer Learning Library\footnote{https://github.com/thuml/Transfer-Learning-Library} for implementation and using the same training parameters: SGD with a learning rate of 0.002, a batch size of 32. For DomainNet, we follow SSRT \cite{sun2022safe} with the same training parameters (SGD with a learning rate of 0.004, a batch size of 32). We used ViT-B/16 \cite{dosovitskiy2020vit} pretrained on ImageNet \cite{deng2009imagenet} as the backbone across all experiments. 


\subsection{Results}
As shown in Tables \ref{tab:visda} to \ref{tab:home}, CAT achieves significant and consistent performance improvements for existing adversarial UDA approaches over various datasets and domains.

\textbf{VisDA-2017}: For the VisDA-2017 dataset, we add our proposed adversarial training plug-in, CAT, to existing adversarial UDA algorithms including CDAN, MCC, SDAT, and MIC and show the results in Table \ref{tab:visda}, where green numbers in the buckets suggest out-perform while red numbers suggest under-perform. We can see that the addition of CAT improves the performance on all baseline adversrial UDA approaches across majority of source-to-target adaptation scenarios. MIC+CAT improves the SoTA adversarial UDA (MIC) performance on the VisDA-2017 by 0.5\% to 93.3\%. Additionally, CAT also enhances the performance of other listed adversarial UDA approaches by a margin from 1.2\% to 4.4\%. This indicates that the proposed CAT can consistently and effectively improve the adversarial training and increase the overall model performance.

\begin{table*}[]
\centering
\small
\setlength{\tabcolsep}{4pt}
\setlength{\aboverulesep}{0pt}
\setlength{\belowrulesep}{0pt}
\caption{Image classification acc. in \% on VisDA-2017 for UDA.}
    \label{tab:visda}
\begin{tabular}{c|cccccccccccc|c}
\toprule
{Method}  & Pln & Bik & Bus  & Car & Hrs & Knf & Mcy & Per & Plt & Skb & Trn & Trk & Avg. \\
\hline\hline
TVT     & 92.9  & 85.6  & 77.5 & 60.5  & 93.6  & 98.2  & 89.3  & 76.4   & 93.6  & 92   & 91.7  & 55.7  & 83.9 \\
CDTrans & 97.1  & 90.5  & 824  & 77.5  & 96.6   & 96.1  & 93.6  & 88.6   & 97.9  & 869  & 90.3  & 62.8  & 88.4 \\
PMTrans & 98.9 & 93.7 & 84.5 & 73.3 & 99.0 & 98.0 & 96.2 & 67.8 & 94.2 & 98.4 & 96.6 & 49.0 & 87.5\\
SSRT&98.9 & 87.6 & 89.1 & 84.8 & 98.3 & 98.7 & 96.3 & 81.1 & 94.8 & 97.9 & 94.5 & 43.1 & 88.8\\
\midrule\midrule
CDAN &94.3 & 53.0 & 75.7 & 60.5 & 93.9 & 98.3 & {96.4} & 77.5 & 91.6 & 81.8 & 87.4 & 45.2 & 79.6\\
\textbf{w. CAT} & {96.2}  & {81.0} & {83.8} & 55.2& 91.7& {97.6}&93.3 &{78.2} &{95.5} &{93.0} & {90.3}& {52.0}& 84.0 \textcolor{dark_green}{(4.4)$\uparrow$} 
\\\midrule

MCC & 96.2 &80.2 &78.5 &68.9 &96.9 &98.9 &94.1 &79.8 &96.1 &94.7 &92.4 &60.5    & 86.4 \\ 
\textbf{w. CAT} & {97.9} &{88.3} &{79.9} &{85.3} &{97.4} &96.5 &90.2 &{82.1} &95.1 &{97.2} &{92.6} &58.1    & 88.3 \textcolor{dark_green}{(1.9)$\uparrow$}\\ 
\midrule
SDAT    & 98.4  & 90.9  & 85.4 & 82.1  & 98.5   & 97.6  & 96.3  & 86.1  & 96.2  & 96.7 & {92.9}  & 56.8   & 89.8 \\
\textbf{w. CAT} & {98.4}   & {92.3}  & {88.2} & {82.8}  & {98.6}  & {98.6}    & 94.4  & {86.5}  & {97.8}  & {98.4} & 92.7  & {62.7}   & {91.0} \textcolor{dark_green}{(1.2)$\uparrow$}\\ 
\midrule
MIC     & {99.0}    & 93.3  & 86.5 & \textbf{87.6}  & 98.9  & 99.0    & \textbf{97.2}  & 89.8  & 98.9  & 98.9 & \textbf{96.5}  & 68.0    & 92.8\\
\textbf{w. CAT} & \textbf{99.0}    & \textbf{94.3}  & \textbf{90.1} & 87.0  & \textbf{99.0}  & \textbf{99.5}   & 95.5  & \textbf{90.4}  & \textbf{98.9}  & \textbf{98.9} & 95.8  & \textbf{71.1}   & \textbf{93.3} \textcolor{dark_green}{(0.5)$\uparrow$}\\ 
\bottomrule
\end{tabular}
\end{table*}

\textbf{DomainNet}: Table \ref{tab:domainnet} shows the experiment results on the largest and most challenging DomainNet dataset across five domains. We conducted a comprehensive experiment over several adversarial UDA approaches and reported their performance with the proposed CAT plug-in. Overall, CAT improves nearly all of the adaptation scenarios significantly with only two minor under-perform (-0.1\% and -0.2\%). Specifically, with the help of CAT, the average performance of the SoTA algorithm, SSRT, is improved by 1.0\% to 58.0\%, which is a significant lift considering the large number of classes (345 classes) and images present in DomainNet. The performance of DANN and SDAT is also improved by 1.6\% and 2.7\%, respectively. On specific source-to-target adaptation scenarios like inf $\rightarrow$ skt, the SSRT+CAT performance increase is up to 5.1\%. 
\begin{table}[h]
\centering
\scriptsize
\setlength{\tabcolsep}{4pt}
\setlength{\aboverulesep}{0pt}
\setlength{\belowrulesep}{0pt}
\caption{Image classification acc. in \% on DomainNet for UDA.}\label{tab:domainnet}
\begin{tabular}{c|cccccc|c|cccccc}
\toprule
\multirow{ 2}{*}{DANN} & \multirow{ 2}{*}{clp}  & \multirow{ 2}{*}{inf}  & \multirow{ 2}{*}{pnt}  & \multirow{ 2}{*}{rel}  &\multirow{ 2}{*}{ skt}  & \multirow{ 2}{*}{Avg.} & DANN & \multirow{ 2}{*}{clp}  & \multirow{ 2}{*}{inf}  & \multirow{ 2}{*}{pnt}  & \multirow{ 2}{*}{rel}  & \multirow{ 2}{*}{skt}  & \multirow{ 2}{*}{Avg}  \\
 &  &  &  &  &  &  & +CAT &  &  &  &  &  & \\
\midrule
\multirow{ 2}{*}{clp}  & -    & 30.9 & 53.3 & 72.7 & 55.4 & 53.1 & \multirow{ 2}{*}{clp}      & -    & 31.5 & 55.6 & 73.6 & 57.0 & 54.4 \\
     &      &      &      &      &      &      &          & -    & \textcolor{dark_green}{(0.6)}  & \textcolor{dark_green}{(2.3)}  & \textcolor{dark_green}{(0.9)}  & \textcolor{dark_green}{(1.6)}  & \textcolor{dark_green}{(1.4)}  \\
\multirow{ 2}{*}{inf}  & 43   & -    & 40.8 & 56.4 & 35.9 & 44.0 & \multirow{ 2}{*}{inf}      & 47.3 & -    & 45.6 & 57.9 & 40.7 & 47.9 \\
     &      &      &      &      &      &      &          & \textcolor{dark_green}{(4.3)}  & -    & \textcolor{dark_green}{(4.8)}  & \textcolor{dark_green}{(1.5)}  & \textcolor{dark_green}{(4.8)}  & \textcolor{dark_green}{(3.9)} \\
\multirow{ 2}{*}{pnt}  & 55.7 & 28.6 & -    & 70.5 & 48.3 & 50.8 & \multirow{ 2}{*}{pnt}      & 56.6 & 29.4 & -    & 70.5 & 49.8 & 51.6 \\
     &      &      &      &      &      &      &          & \textcolor{dark_green}{(0.9)}  & \textcolor{dark_green}{(0.8)}  & -    & \textcolor{dark_green}{(0.0)}  & \textcolor{dark_green}{(1.5)}  & \textcolor{dark_green}{(0.8)}  \\
\multirow{ 2}{*}{rel}  & 62.3 & 32.5 & 62.5 & -    & 50.7 & 52.0 & \multirow{ 2}{*}{rel}      & 64.1 & 33.9 & 63.1 & -    & 51.8 & 53.2 \\
     &      &      &      &      &      &      &          & \textcolor{dark_green}{(1.8)}  & \textcolor{dark_green}{(1.4)}  & \textcolor{dark_green}{(0.6)}  & -    & \textcolor{dark_green}{(1.1)}  & \textcolor{dark_green}{(1.2)}  \\
\multirow{ 2}{*}{skt}  & 66.4 & 30.6 & 58   & 70.1 & -    & 56.3 & \multirow{ 2}{*}{skt}      & 67.0 & 31.1 & 59.1 & 70.1 & -    & 56.8 \\
     &      &      &      &      &      &      &          & \textcolor{dark_green}{(0.6)}  & \textcolor{dark_green}{(0.5)}  & \textcolor{dark_green}{(1.1)}  & \textcolor{dark_green}{(0.0)}  & -    & \textcolor{dark_green}{(0.5)}  \\
\multirow{ 2}{*}{Avg.} & 56.9 & 30.7 & 53.7 & 67.4 & 47.6 & 51.2 & \multirow{ 2}{*}{Avg.}      & 58.8 & 31.5 & 55.9 & 68.0 & 49.8 & 52.8 \\
     &      &      &      &      &      &      &          & \textcolor{dark_green}{(1.9)}  & \textcolor{dark_green}{(0.8)}  & \textcolor{dark_green}{(2.2)}  & \textcolor{dark_green}{(0.6)}  & \textcolor{dark_green}{(2.3)}  & \textcolor{dark_green}{(1.6)}  \\\midrule\midrule
\multirow{ 2}{*}{SDAT} &\multirow{ 2}{*}{ clp}  &\multirow{ 2}{*}{ inf } &\multirow{ 2}{*}{ pnt}  &\multirow{ 2}{*}{ rel}  &\multirow{ 2}{*}{ skt}  &\multirow{ 2}{*}{ Avg.} & SDAT & \multirow{ 2}{*}{clp}  & \multirow{ 2}{*}{inf}  & \multirow{ 2}{*}{pnt}  & \multirow{ 2}{*}{rel}  & \multirow{ 2}{*}{skt}  & \multirow{ 2}{*}{Avg.} \\
 &  &  &  &  &  &  & +CAT &  &  &  &  &  & \\\midrule
\multirow{ 2}{*}{clp}  & -    & 28.2 & 51.5 & 68.6 & 55.9 & 51.1 & \multirow{ 2}{*}{clp}      & -    & 31.5 & 55.5 & 72.7 & 58.5 & 54.6 \\
     &      &      &      &      &      &      &          & -    & \textcolor{dark_green}{(3.3)}  & \textcolor{dark_green}{(4.0)}  & \textcolor{dark_green}{(4.1)}  & \textcolor{dark_green}{(2.6)}  & \textcolor{dark_green}{(3.5)}  \\
\multirow{ 2}{*}{inf}  & 40.3 & -    & 41.7 & 53.9 & 35.4 & 42.8 & \multirow{ 2}{*}{inf}      & 43.7 & -    & 45.4 & 57.6 & 37.8 & 46.1 \\
     &      &      &      &      &      &      &          & \textcolor{dark_green}{(3.4)}  & -    & \textcolor{dark_green}{(3.7)}  & \textcolor{dark_green}{(3.7)}  & \textcolor{dark_green}{(2.4)}  & \textcolor{dark_green}{(3.3)}  \\
\multirow{ 2}{*}{pnt}  & 50.9 & 27.5 & -    & 64.8 & 45.3 & 47.1 & \multirow{ 2}{*}{pnt}      & 53.4 & 28.4 & -    & 67.0 & 47.3 & 49.0 \\
     &      &      &      &      &      &      &          & \textcolor{dark_green}{(2.5)}  & \textcolor{dark_green}{(0.9)}  & -    & \textcolor{dark_green}{(2.2)}  & \textcolor{dark_green}{(2.0)}  & \textcolor{dark_green}{(1.9 )} \\
\multirow{ 2}{*}{rel}  & 59.8 & 32.1 & 61.3 & -    & 49.0 & 50.6 & \multirow{ 2}{*}{rel}      & 62.7 & 33.2 & 63.4 & -    & 50.7 & 52.5 \\
     &      &      &      &      &      &      &          & \textcolor{dark_green}{(2.9)}  & \textcolor{dark_green}{(1.1)}  & \textcolor{dark_green}{(2.1)}  & -    & \textcolor{dark_green}{(1.7)}  & \textcolor{dark_green}{(2.0)}  \\
\multirow{ 2}{*}{skt}  & 62.3 & 29.2 & 53.6 & 62.8 & -    & 52.0 & \multirow{ 2}{*}{skt}      & 65.6 & 30.7 & 56.8 & 65.2 & -    & 54.6 \\
     &      &      &      &      &      &      &          & \textcolor{dark_green}{(3.3)}  & \textcolor{dark_green}{(1.5)}  & \textcolor{dark_green}{(3.2)}  & \textcolor{dark_green}{(2.4 )} & -    & \textcolor{dark_green}{(2.6)}  \\
\multirow{ 2}{*}{Avg.} & 53.3 & 29.3 & 52.0 & 62.5 & 46.4 & 48.7 & \multirow{ 2}{*}{Avg.}     & 56.4 & 31.0 & 55.3 & 65.6 & 48.6 & 51.4 \\
     &      &      &      &      &      &      &          & \textcolor{dark_green}{(3.1)}  & \textcolor{dark_green}{(1.7)}  & \textcolor{dark_green}{(3.3 )} & \textcolor{dark_green}{(3.1 )} &\textcolor{dark_green}{( 2.2)}  & \textcolor{dark_green}{(2.7)}  \\\midrule\midrule
\multirow{ 2}{*}{SSRT} & \multirow{ 2}{*}{clp}  & \multirow{ 2}{*}{inf}  & \multirow{ 2}{*}{pnt}  & \multirow{ 2}{*}{rel}  & \multirow{ 2}{*}{skt}  & \multirow{ 2}{*}{Avg.} & SSRT & \multirow{ 2}{*}{clp}  & \multirow{ 2}{*}{inf}  & \multirow{ 2}{*}{pnt}  & \multirow{ 2}{*}{rel}  & \multirow{ 2}{*}{skt}  & \multirow{ 2}{*}{Avg.} \\
 &  &  &  &  &  &  & +CAT &  &  &  &  &  & \\ \midrule
\multirow{ 2}{*}{clp}  & -    & 33.8 & 60.2 & 75.8 & 59.8 & 57.4 & \multirow{ 2}{*}{clp}      & -    & 34.6 & 60.7 & 75.7 & 60.6 & 57.9 \\
     &      &      &      &      &      &      &          & -   & \textcolor{dark_green}{(0.8 )} & \textcolor{dark_green}{(0.5 )} & \textcolor{dark_red}{(-0.1)} & \textcolor{dark_green}{(0.8 )} & \textcolor{dark_green}{(0.5)}   \\
\multirow{ 2}{*}{inf}  & 55.5 & -    & 54.0 & 68.2 & 44.7 & 55.6 & \multirow{ 2}{*}{inf}      & 57.3 & -    & 55.7 & 68   & 49.8 & 57.7 \\
     &      &      &      &      &      &      &          & \textcolor{dark_green}{(1.8 )} & -    & \textcolor{dark_green}{(1.7 )} & \textcolor{dark_red}{(-0.2)} & \textcolor{dark_green}{(5.1 )} & \textcolor{dark_green}{(2.1 )} \\
\multirow{ 2}{*}{pnt}  & 61.7 & 28.5 & -    & 71.4 & 55.2 & 54.2 & \multirow{ 2}{*}{pnt}      & 62.2 & 31.0 & -    & 71.9 & 56.1 & 55.3 \\
     &      &      &      &      &      &      &          & \textcolor{dark_green}{(0.5 )} & \textcolor{dark_green}{(2.5 )} & -    & \textcolor{dark_green}{(0.5 )} & \textcolor{dark_green}{(0.9 )} & \textcolor{dark_green}{(1.1 )} \\
\multirow{ 2}{*}{rel}  & 69.9 & 37.1 & 66.0 & -    & 58.9 & 58.0 & \multirow{ 2}{*}{rel}      & 70.9 & 37.2 & 66.4 & -    & 60.1 & 58.7 \\
     &      &      &      &      &      &      &          & \textcolor{dark_green}{(1.0 )} & \textcolor{dark_green}{(0.1 )} & \textcolor{dark_green}{(0.4 )} & -    & \textcolor{dark_green}{(1.2 )} & \textcolor{dark_green}{(0.7)}  \\
\multirow{ 2}{*}{skt}  & 70.6 & 32.8 & 62.2 & 73.2 & -    & 59.7 & \multirow{ 2}{*}{skt}      & 71.4 & 32.9 & 63.3 & 73.4 & -    & 60.3 \\
     &      &      &      &      &      &      &         & \textcolor{dark_green}{(0.8 )} & \textcolor{dark_green}{(0.1 )} & \textcolor{dark_green}{(1.1 )} & \textcolor{dark_green}{(0.2)}   & -   & \textcolor{dark_green}{(0.6)}   \\
\multirow{ 2}{*}{Avg.} & 64.4 & 33.1 & 60.6 & 72.2 & 54.7 & 57.0 & \multirow{ 2}{*}{Avg.}     & 65.5 & 33.9 & 61.5 & 72.3 & 56.7 & 58.0 \\
     &      &      &      &      &      &      &          & \textcolor{dark_green}{(1.1 )} & \textcolor{dark_green}{(0.8 )} & \textcolor{dark_green}{(0.9 )} & \textcolor{dark_green}{(0.1 )} & \textcolor{dark_green}{(2.0 )} & \textcolor{dark_green}{(1.0 )}\\ \bottomrule
\end{tabular}

\end{table}

\textbf{Office-Home}: As seen in Table \ref{tab:home} - the quantitative results on Office-Home dataset, our proposed CAT plug-in achieves notable performance gains on CDAN, MCC, and SDAT approach, as well as other Vit-based approaches. When comparing the performance improvement across different domain datasets, CAT particularly gains bigger performance improvement for the domain adaptation scenarios such as AC, PC, RC that have exhibits less promising results by the base algorithms. When comparing the performance improvement across different base algorithms, the CAT plug-in obtains bigger performance improvement over less complex algorithms, indicating CAT's effectiveness on increasing the difficulty for the discriminator. Overall, the results show that our proposed method, CAT, can explicitly generalize domain invariant features by improving the domain adversarial training process.

\begin{table*}[t]
\centering
\setlength{\tabcolsep}{4pt}
\setlength{\aboverulesep}{0pt}
\setlength{\belowrulesep}{0pt}
\caption{Image classification acc. in \% on Office-Home for UDA.}\label{tab:home}
\begin{tabular}{c|ccccccccccccc}
\toprule
Method  & AC   & AP   & AR   & CA   & CP   & CR   & PA   & PC   & PR   & RA   & RC   & RP   & Avg.  \\ \hline\hline
CDTrans & 68.8 & 85.0   & 86.9 & 81.5 & 87.1 & 87.3 & 79.6 & 63.3 & 88.2 & 82   & 66   & 90.6 & 80.5 \\
TVT     & 74.9 & 86.8 & 89.5 & 82.8 & 87.9 & 88.3 & 79.8 & 71.9 & 90.1 & 85.5 & 74.6 & 90.6 & 83.6 \\ \midrule\midrule
CDAN&62.6 & 82.9 & 87.2 & 79.2 & 84.9 & 87.1 & 77.9 & 63.3 & 88.7 & 83.1 & 63.5 & 90.8 & 79.3\\
\textbf{w. CAT}& 65.1 &  84.3  & 88.2 & 82.6 & 85.0 & 87.5 & 81.2 & 64.6 & 89.0 & 85.3 & 69.0 & 91.2 & 81.1 \textcolor{dark_green}{(1.8)$\uparrow$}\\ \midrule
MCC & 67.0 & 84.8 & 90.2 & 83.4 & 87.3 & 89.3 & 80.7 & 64.4 & 90.0 & 86.6 & 70.4 & 91.9 & 82.2\\
\textbf{w. CAT} & 67.1 &  \textbf{88.9}  & 89.7 & 84.1 & \textbf{87.4} & 89.5 & 81.8 & 64.6 & 90.2& 85.2& 70.1 &91.3  & 82.5 \textcolor{dark_green}{(0.3)$\uparrow$}\\\midrule
SDAT    & 70.8 & 87.0   & \textbf{90.5} & 85.2 & 87.3 & \textbf{89.7} & \textbf{84.1} & 70.7 & \textbf{90.6} & \textbf{88.3} & 75.5 & 92.1 & 84.3 \\
\textbf{w. CAT} & \textbf{75.6} & 87.5 & 90.4 & \textbf{87.7} & 88.6   & 89.6 & 83.5 & \textbf{72.8} & 90.2 & 87.0 & \textbf{76.4} & \textbf{92.4} & \textbf{85.1} \textcolor{dark_green}{(0.8)$\uparrow$}\\
\bottomrule
\end{tabular}
\end{table*}

\section{Ablation Study and Discussions} 
~~~~\textbf{Effectiveness of CAT}: As shown in Tables \ref{tab:visda}-\ref{tab:home}, existing adversarial UDA approaches received substantial performance improvements by integrating with CAT in all datasets. Especially in DomainNet dataset, which is the largest UDA dataset to date, CAT improves 58 out of 60 compared algorithm and adaptation scenario combinations. It shows the effectiveness of increasing the competitiveness of adversarial training.

Additionally, to validate the effectiveness of CAT's feature alignment capability, we compare its performance with another feature alignment approach, KL-divergence (KLD), in Table \ref{tab:cat-vs-kld}. The results for CDAN and MIC show that using KLD can slightly increase the performance for the original models but less effective comparing to CAT, indicating that the adversarial training is improved due to the increased difficulty for domain classification (i.e., the discriminator). However, for MCC and SDAT, due to the unpaired issue of unlabeled data samples, samples from different class were aligned, which damages the their generalization and the performance of the classifier also decreases. 

\begin{table}[h]
\centering
\setlength{\tabcolsep}{4pt}
\setlength{\aboverulesep}{0pt}
\setlength{\belowrulesep}{0pt}
\caption{Effective of CAT vs KLD.} \label{tab:cat-vs-kld}
\begin{tabular}{c|cccc}
\toprule
         & CDAN & MCC  & SDAT & MIC  \\ \midrule
Baseline & 79.6 & 86.4 & 89.8 & 92.8 \\
+ KLD   & 81.1 & 86.1 & 89.5 & 92.9 \\
+ CAT   & 84.0 & 88.3 & 91.0 & 93.3 \\ \bottomrule
\end{tabular}
\end{table}

\begin{figure}[h]
  \centering
  \includegraphics[width=0.8\linewidth]{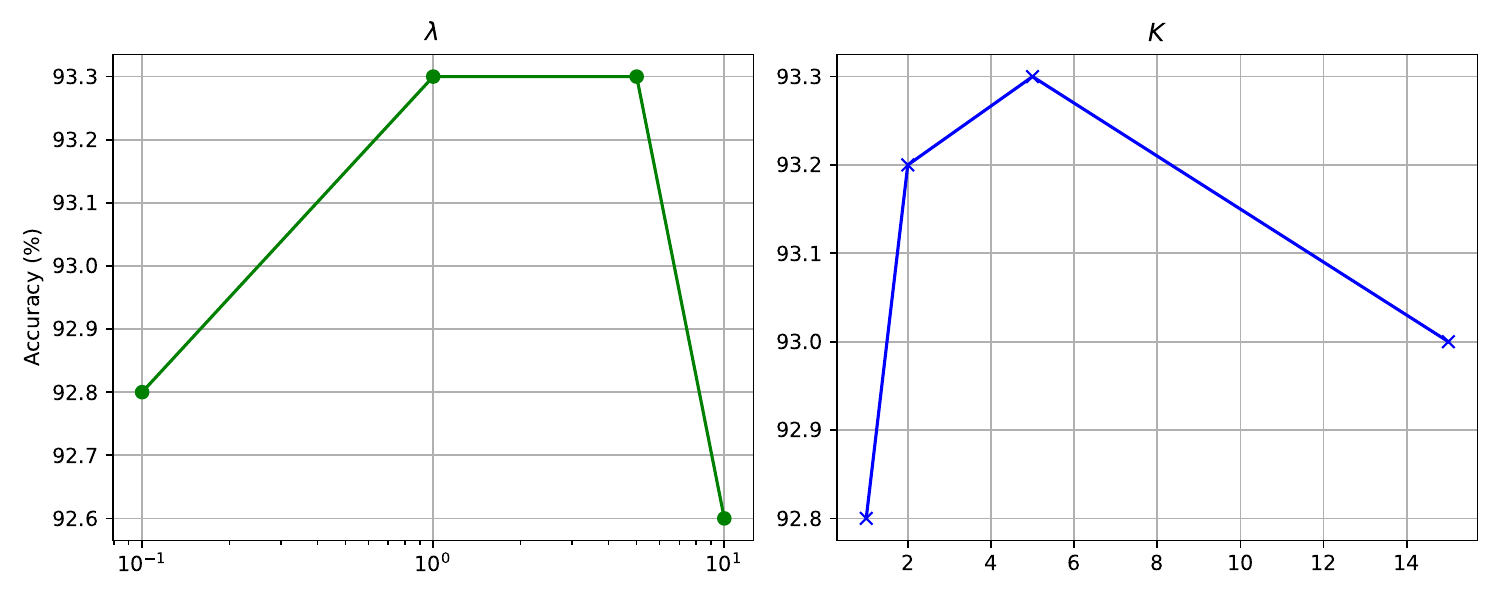}
  \caption{Impact of contrastive coefficient $\lambda$ and neighbor size.}\label{fig:abl}
\end{figure}

\textbf{Impact of contrastive coefficient $\lambda$ and neighbor size $K$}: We study the impact of hyperparameters of CAT, $\lambda$ and $K$, on VisDA-2017 with MIC backbone. We vary $\lambda$ in $[0.1, 1,5,10]$ and $K$ in $[1,2,5,15]$. The results are presented in Fig. \ref{fig:abl}. Generally, our model is less sensitive to the change of $\lambda$, where the model performance is peaked at 93.3 when we choose $\lambda = 1, 5$. The performance of CAT continuously outperforms the baseline method before $\lambda$ become too large. The reason is that a large $\lambda$ may let $\mathcal{L}_{con}$ dominate the loss function and make the model only focus of source-to-target alignment without making accurate source domain classification and domain discrimination. As for the impact of neighbor size $K$, the model is more sensitive to it, where the best model performance 93.3 is achieved with $K = 5$. Generally, a larger $K$ leads to a better model performance, indicating the contrastive learning between the source anchor and similar target samples is more effective when we select more samples from the target domain. However, when the number of neighbors is too large, noisy samples from other classes might be introduced, hence the performance starts to drop.

\begin{figure}[t]
\centering
\begin{subfigure}{.35\textwidth}
  \centering
  \includegraphics[width=1\linewidth]{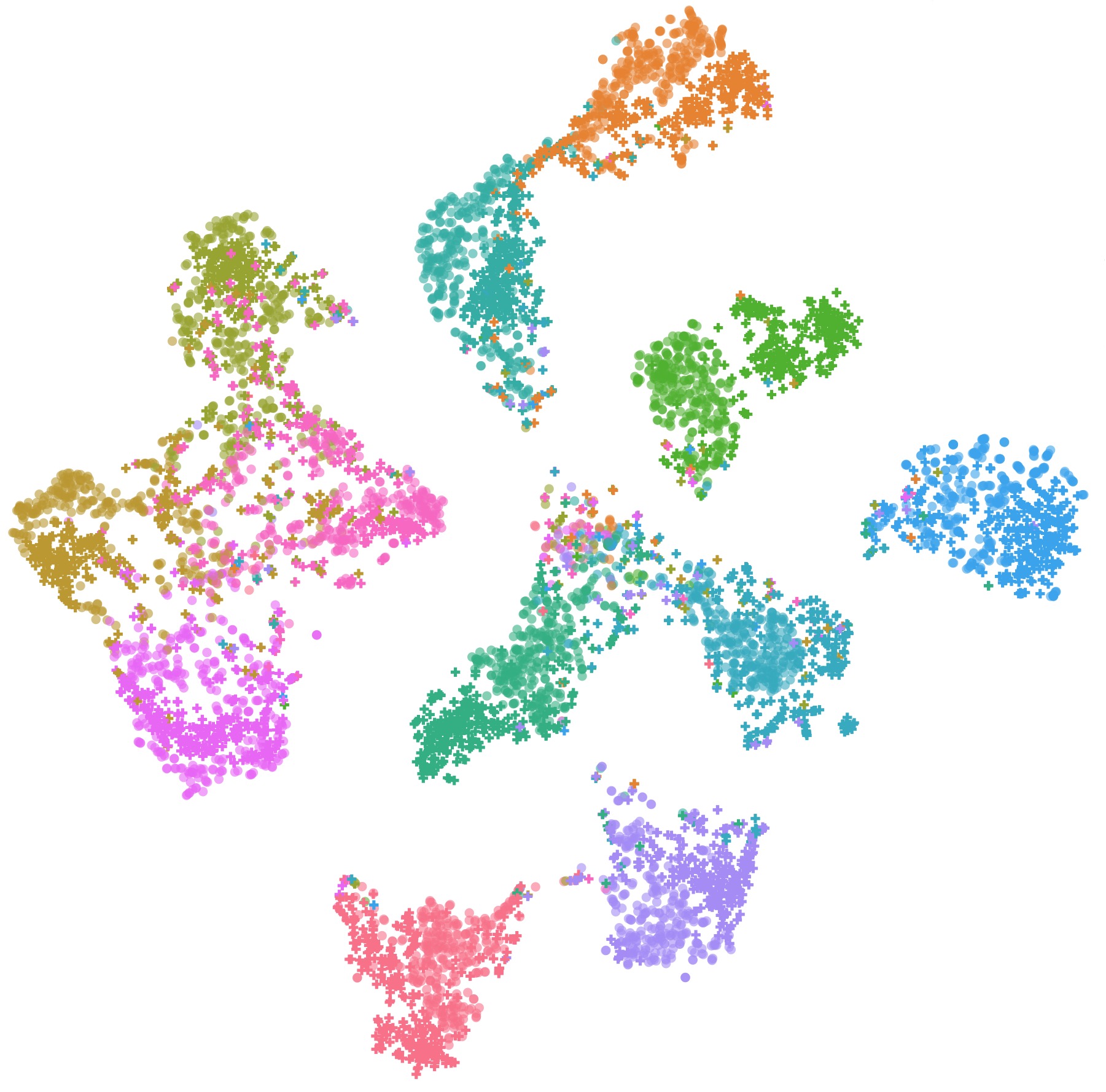}
  \caption{SDAT}
\end{subfigure}
\begin{subfigure}{.35\textwidth}
  \centering
  \includegraphics[width=1\linewidth]{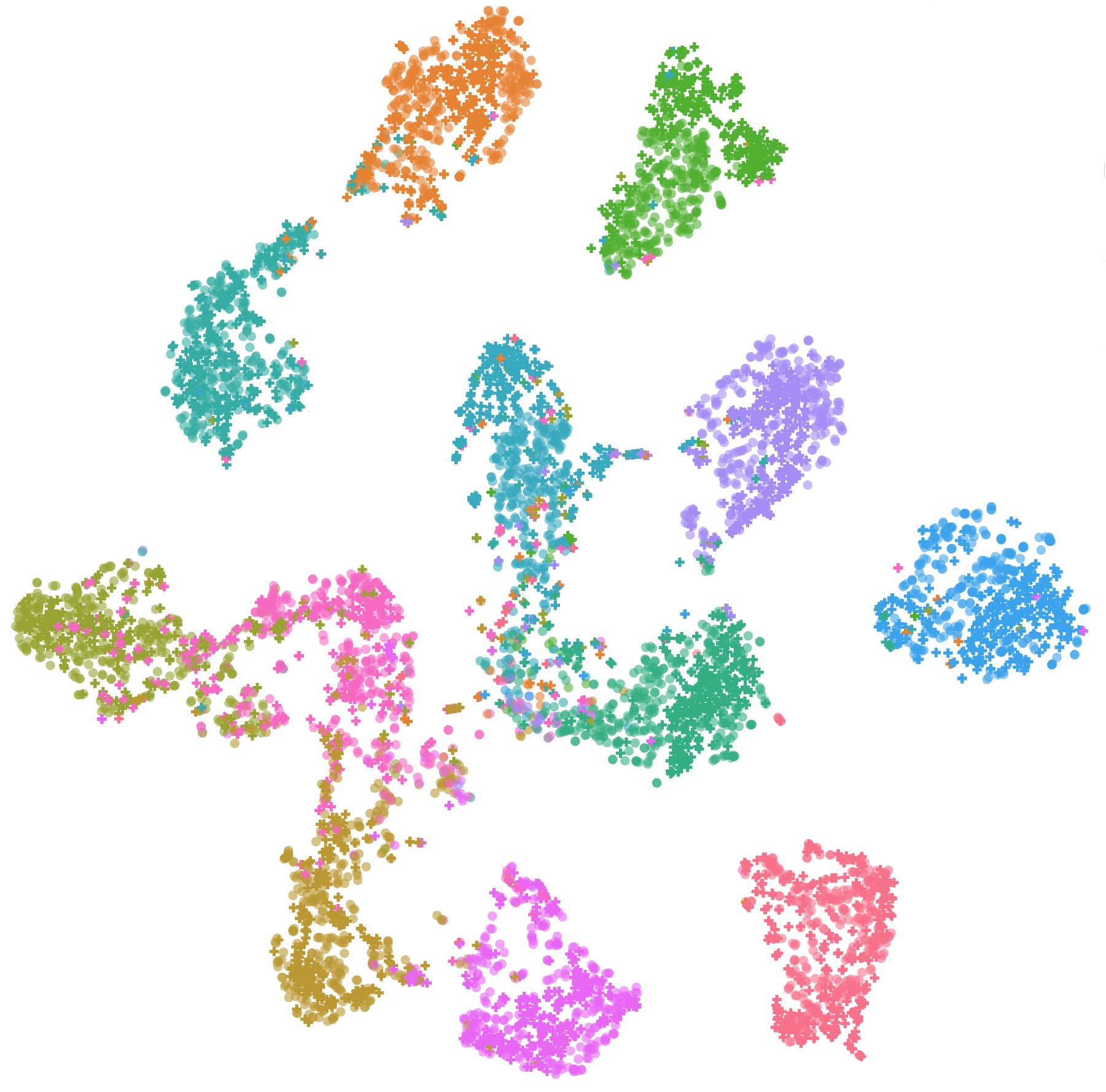}
  \caption{SDAT+CAT}
\end{subfigure}

\begin{subfigure}{.35\textwidth}
  \centering
  \includegraphics[width=1\linewidth]{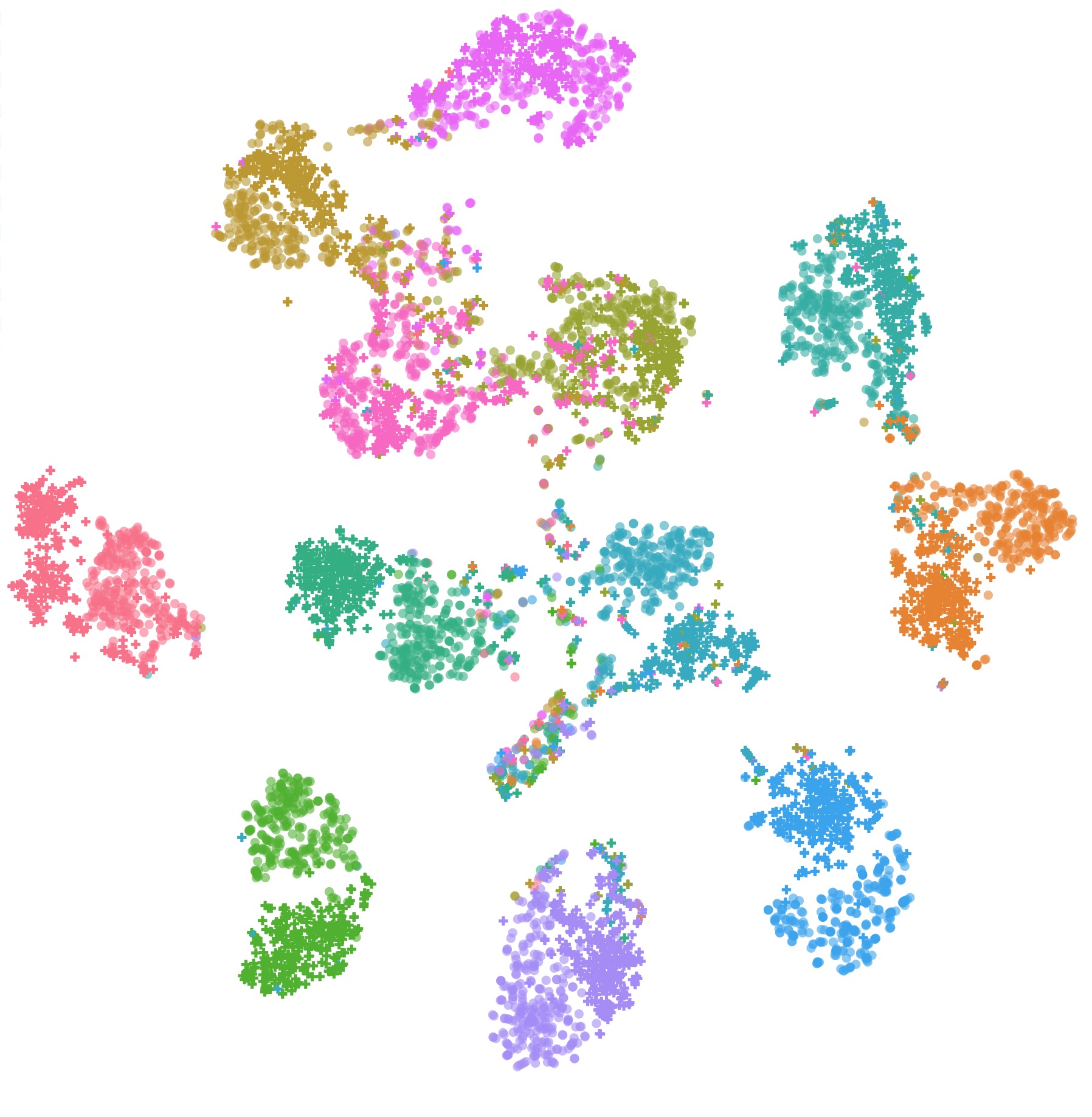}
  \caption{MIC}
\end{subfigure}
\begin{subfigure}{.35\textwidth}
  \centering
  \includegraphics[width=1\linewidth]{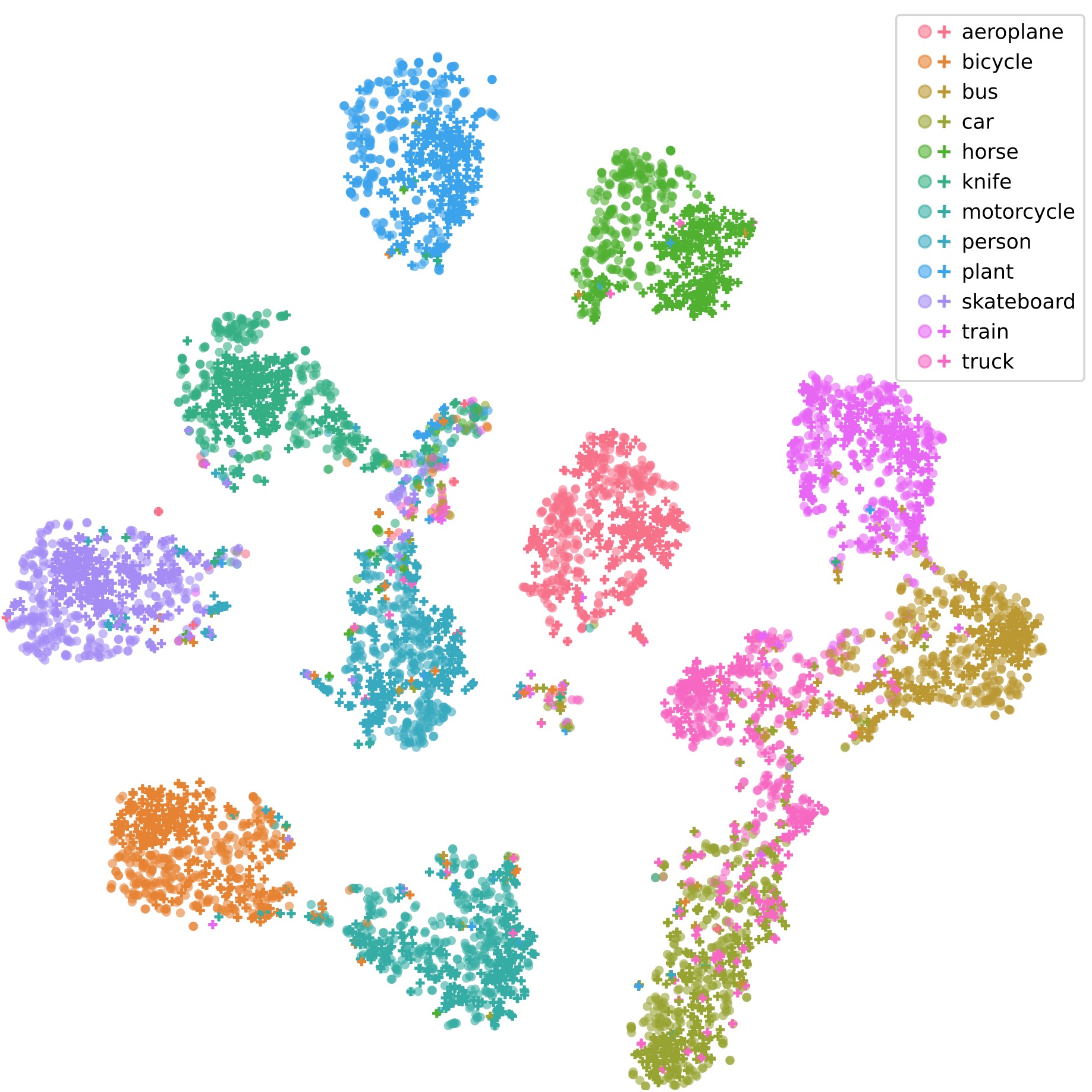}
  \caption{MIC+CAT}
\end{subfigure}
\caption{Visualization with t-SNE for adaptation methods w/wo CAT (best viewed in color). }
\label{fig:tsne}
\end{figure}

\textbf{Visualization}: We compare our proposed method, CAT, with the model trained without it to verify the effectiveness of introducing the contrastive adversarial training. The visualization results using t-SNE are shown in Fig. \ref{fig:tsne}, where dots denotes samples from source domain, and + denotes the ones in target domain. In this visualization, 300 samples were randomly selected from each class in the VisDA-2017 dataset. We can clearly see that the target features and the source features of the same class are fully overlapped and collapse into one another, which indicates a desirable alignment between source and target domains. In contrast, for the methods without CAT, although the samples from the same class but different domains are also mapped into neighboring region, there is a clear boundary between the two clusters of samples that are from different domains, which shows a less effective feature alignments between source and target domains. The comparison suggests that the proposed CAT can explicitly reduce the bias towards the labeled source domain and improve target-to-source feature alignment, which attributes to the improved domain adaptation results shown in Fig. \ref{fig:tsne}.

Additionally, compared to Fig. \ref{fig:tsne} (a) and (c), the features from different classes in Fig. \ref{fig:tsne} (b) and (d) are better separated from each other, especially for classes with similar objects such as Car, Truck, Bus, and Train. This suggests that the proposed CAT can effectively encourage interclass separation by leverage the contrastive training, leading to a easier decision boundary for the classifier.

\textbf{Effectiveness of cross-domain intraclass alignment}: We also calculate the $\mathcal{A}$-distance \cite{ben2006analysis} to quantitatively measure the domain distribution discrepancy for the data samples used in t-SNE visualization:
\begin{equation}
    dist_\mathcal{A} = 2 (1-2\epsilon),
\end{equation}
where $\epsilon$ is the test error of a binary classifier trained 10 epochs to discriminate the source samples from the target samples. The results show that before applying CAT, the $\mathcal{A}$-distance is 1.84 and 1.77 (lower the better) for SDAT and MIC, respectively. In contrast, the $\mathcal{A}$-distance for SDAT+CAT (1.30) and MIC (1.29) is signifcantly lower, suggesting a better domain alignment.

\begin{figure}
\centering
\makebox[10pt]{\raisebox{30pt}{\rotatebox[origin=c]{90}{Source}}}%
\begin{subfigure}{.3\textwidth}
  \centering
  \includegraphics[width=1\linewidth, height=2.5cm]{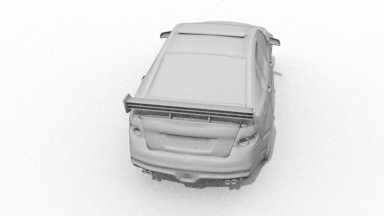}
  \caption{Common car}
\end{subfigure}
\begin{subfigure}{.3\textwidth}
  \centering
  \includegraphics[width=1\linewidth, height=2.5cm]{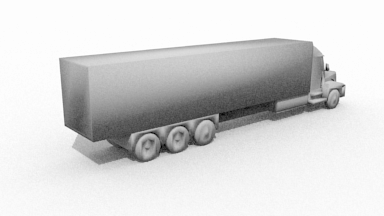}
  \caption{Common truck}
\end{subfigure}
\begin{subfigure}{.3\textwidth}
  \centering
  \includegraphics[width=1\linewidth, height=2.5cm]{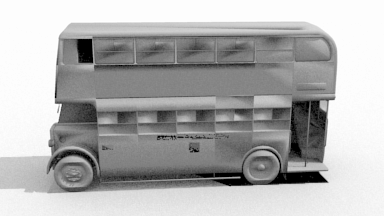}
  \caption{Common bus}
\end{subfigure}

\makebox[10pt]{\raisebox{30pt}{\rotatebox[origin=c]{90}{Target}}}%
\begin{subfigure}{.3\textwidth}
  \centering
  \includegraphics[width=1\linewidth, height=2.5cm]{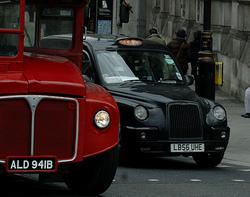}
  \caption{Special car}
\end{subfigure}
\begin{subfigure}{.3\textwidth}
  \centering
  \includegraphics[width=1\linewidth, height=2.5cm]{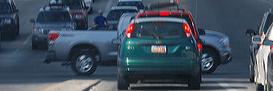}
  \caption{Special truck}
\end{subfigure}
\begin{subfigure}{.3\textwidth}
  \centering
  \includegraphics[width=1\linewidth, height=2.5cm]{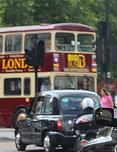}
  \caption{Special bus}
\end{subfigure}
\caption{Hard to classify cases in DomainNet.}
\label{fig:hard-case}
\end{figure}

\textbf{Limitations of using Contrastive Training}: From Fig. \ref{fig:tsne}, we observe that many data samples from different classes are closely clustered in the t-SNE visualization, especially in classes Car, Truck, Bus. Some samples are even clustered to wrong classes after contrastive training. To find the root cause of such deficiency for our proposed contrastive training method, the raw images from the target domain of the VisDA-2017 dataset are illustrated in Fig. \ref{fig:hard-case}, where the first row shows the common objects in these classes from the source domain while the second row shows some special objects from the target domain. As shown in Fig. \ref{fig:hard-case} (d)-(f), there are some special images that contain the objects from other classes and these objects even dominate the areas of the particular images. This is because most of the target domain images for these classes were taken from street, which does not guarantee that a single object is contained in each image. Hence, during the contrastive training, the non-target objects in such images lead to the images being mapped to the wrong classes (typically, the classes that are related to the non-target objects). Nevertheless, such special adaptation cases further validate the effectiveness of CAT in pulling similar target samples to the source domain anchors, while pushing the dis-similar target domain samples away from the anchors.


\section{Conclusions}
In this paper, we presented Contrastive Adversarial Training (CAT) for unsupervised domain adaptation, a plug-in module to improve the learning of domain invariant feature by increasing the competitiveness during adversarial training. Through increasing the difficulty of the discriminator for domain classification, CAT forces the feature extractor to generate more robust domain invariant features. In the meantime, thanking to the target sample features being clustered closely with their source anchor samples, the classifier trained on the labeled source domain data can be easily adopted into the unlabelled target domain. The comprehensive experiments shown that CAT can effectively align the source and target feature distributions, and substantially improve the performance of existing adversarial approaches over large and complex UDA datasets.

\bibliographystyle{splncs04}
\bibliography{main}
\end{document}